\documentclass[11pt]{article}

\usepackage[a4paper,margin=2.5cm]{geometry}
\usepackage{graphicx}
\usepackage{booktabs}
\usepackage{amsmath}
\usepackage{siunitx}
\usepackage[british]{babel}
\usepackage{natbib}
\usepackage[colorlinks=true,linkcolor=blue,citecolor=blue,urlcolor=blue]{hyperref}
\usepackage{authblk}

\newcommand{\kgN}{kg\,N\,ha$^{-1}$}
\newcommand{\tha}{t\,ha$^{-1}$}

\title{Accurate prediction is not profitable advice: profit-based
evaluation of machine learning nitrogen recommendations in winter
wheat}

\author[1]{Xulong Wang}
\author[1]{Po Yang\thanks{Corresponding author: po.yang@sheffield.ac.uk}}
\affil[1]{\small University of Sheffield}
\date{}

\begin{document}
\maketitle

\begin{abstract}
Nitrogen rates for winter wheat are set before the season, under
unknown prices and weather. The standard UK advice does not respond to
prices, yet recent price swings moved the most profitable rate by tens
of kilograms per hectare. Machine learning is often proposed as the
fix. However, it is usually judged on prediction accuracy, and accurate
prediction does not by itself make the recommended rate more
profitable. Our insight is to score nitrogen advice directly by the
profit it forgoes on measured yield response curves. We
build a test bench on 892 such curves from two long running UK
experiments, and sweep the nitrogen to grain price ratio to cover all
price scenarios. On this bench, machine learning fails as a predictor.
No model recovers the best rate within farm tolerance, and the
benchmark noise shows none can. At normal prices, every model also
loses to the standard advice on profit. The gain sits elsewhere. A
simple correction step applied after the model cuts profit losses by a
quarter, while better models and extra features give no gain. The same
frozen correction cuts losses by 43\% at the second site without any
retraining. A hybrid of standard advice plus a damped correction
removes bias and trims rare large losses. The same price sweep also
prices emission cuts, at a cost comparable to current carbon prices.
Machine learning therefore pays as a profit scored correction to
standard advice, not as its replacement.
\end{abstract}

\section{Introduction}\label{sec:intro}

\begin{figure}[t]
  \centering
  \includegraphics[width=\textwidth]{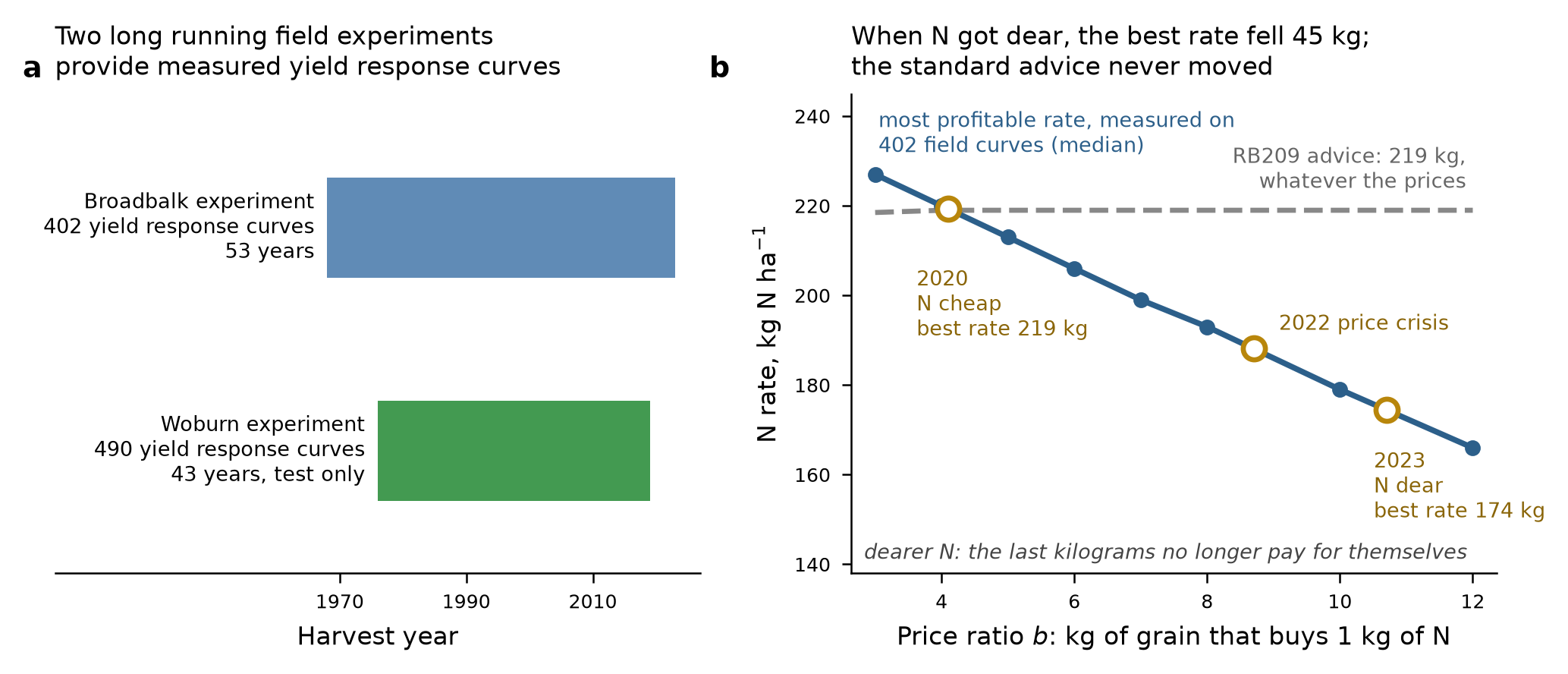}
  \caption{\textbf{Two field experiments; one moving target.}
  \textbf{a}, The two data sources. Broadbalk gives 402 measured yield
  response curves over 53 years. Woburn gives 490 curves over 43 years
  and is used for testing only.
  \textbf{b}, The most profitable N rate, measured on the Broadbalk
  curves (blue, median). It falls as N gets dearer relative to grain.
  Gold rings mark the observed UK price ratios of 2020, 2022, and 2023.
  Those prices alone moved the best rate from 219 to 174~kg. The RB209
  advice (grey) stays at 219~kg throughout.}
  \label{fig:motivation}
\end{figure}

Choosing a nitrogen (N) rate is the largest input decision in UK wheat
growing. Winter wheat covers about 1.6 million hectares, and N
fertiliser is usually the largest purchased input. The rate is
committed before the season, when weather is unknown. Prices are
unknown too, and they moved violently in recent years. Between January
2020 and January 2023, UK ammonium nitrate went from \pounds 234 to a
peak of \pounds 841 per tonne. Feed wheat moved between \pounds 165 and
\pounds 280 per tonne (Table~\ref{tab:banchors}). One number captures
both prices at once. The price ratio $b$ is the grain needed to pay for
one kilogram of N. It went from 4.1 to 10.7 in three years. On measured
yield curves, that move shifts the most profitable rate by about
\SI{45}{\kilo\gram\per\hectare} (Fig.~\ref{fig:motivation}b).

The standard advice cannot follow that moving target. UK growers use
the AHDB Nutrient Management Guide, RB209. It adjusts for soil,
previous crop, and expected yield, but not for prices. In 2022 it
needed an emergency revision to cover crisis prices \citep{ahdb2022}.
That revision is an official admission of the gap.

Machine learning (ML) is the obvious candidate to fill the gap, but it
is judged on the wrong score. Most studies ask whether a model predicts
yield, or the best rate, accurately. Accuracy is then scored in yield
error or in \kgN{} \citep{tanaka}. Neither score measures the
consequence of following the advice. A rate error that costs nothing is
punished. A rare error that costs \pounds 900 per hectare is averaged
away. The decision needs a decision score.

\begin{figure}[t]
  \centering
  \includegraphics[width=\textwidth]{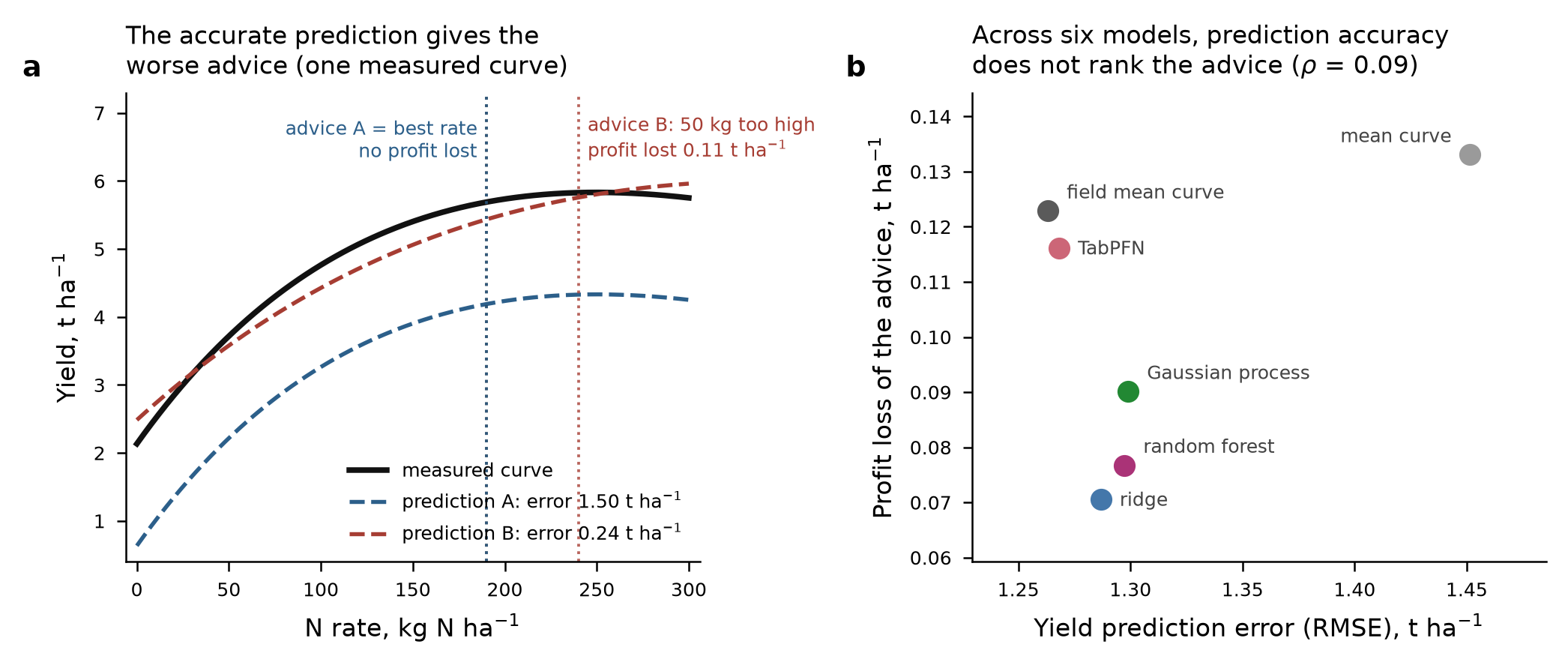}
  \caption{\textbf{Accurate prediction is not profitable advice.}
  \textbf{a}, A measured yield curve from Broadbalk (black) and two
  synthetic predictions. Prediction A is off by 1.50~\tha{} everywhere,
  but it matches the slope, so its advice lands exactly on the best
  rate. Prediction B matches the yields closely (error 0.24~\tha) but
  bends too gently, so its advice is 50~kg too high and loses
  0.11~\tha{} of profit. Prices at $b=5$.
  \textbf{b}, The same decoupling in our six trained models: yield
  prediction error (leave one year out) against the median profit loss
  of the resulting advice at $b=5$. Accuracy does not rank the advice
  (Spearman $\rho = 0.09$).}
  \label{fig:teaser}
\end{figure}

Worse, accurate prediction does not by itself make the advice more
profitable. Figure~\ref{fig:teaser} shows why, on a measured curve from
our data. The best rate depends only on the \emph{slope} of the yield
curve, the point where one more kilogram of N stops paying for itself.
A prediction can be far off in level yet match the slope, and its
advice is then perfect (prediction A). A prediction can match the
yields closely yet bend too gently, and its advice is then 50~kg too
high (prediction B). The same decoupling holds across our six trained
models: their yield accuracy does not rank their advice quality
(Fig.~\ref{fig:teaser}b, Spearman $\rho = 0.09$). Chasing accuracy
therefore optimises the wrong target.

Our insight is simple. Score N advice by the profit it forgoes on
measured yield curves, under every price scenario. Long running field
experiments make this possible. They give hundreds of measured yield
response curves, each with a known best rate at each price ratio. Any
advice can be scored against them, in the unit that matters.

The contributions of this work are summarised as follows:
\begin{itemize}
  \item We propose a profit based evaluation framework for nitrogen
        recommendations. It scores any method by lost profit on 892
        measured yield response curves, and a single sweep of the price
        ratio builds price uncertainty into the score itself
        (\S\ref{sec:framework}).
  \item We show that rate error, the standard score in prior work, is
        ill posed: the measured benchmark carries about 23~\kgN{} of
        noise, so no model can pass a farm tolerance test. We further
        show that on profit, six ML model families all lose to the
        standard advice at normal prices (\S\ref{sec:negative}).
  \item We localise the gain of ML to a post hoc correction step. It
        yields the entire 24\% improvement while model and feature
        changes yield none, and it transfers frozen to a second site
        with a 43\% improvement ($p=0.034$). Building on this, a hybrid
        of standard advice plus a damped ML correction removes bias and
        trims rare large losses (\S\ref{sec:decision},
        \S\ref{sec:external}).
  \item We demonstrate that the price ratio doubles as a carbon price,
        turning the same optimiser into an emission control with a
        worked abatement cost of \pounds 32 to 72 per tonne of CO$_2$e
        (\S\ref{sec:economics}).
\end{itemize}

\section{Data}\label{sec:data}

Two long running UK experiments supply the measured curves. Both
datasets are published open access by the electronic Rothamsted
Archive, e-RA \citep{bbkdata,wrndata}. Broadbalk
at Rothamsted is the oldest fertiliser experiment in the world. Its
plots receive fixed N rates every year. For each field section and
year, we fit a yield response curve through three points. The points
are the yields at 0, 144, and 288~\kgN. The fits share one curvature
parameter and must not bend downwards. Quality screening keeps 402
curves across 53 harvest years, 1968 to 2022. Nine features accompany
each curve, covering weather, previous crop, cultivar era, and field
section. Woburn Ley Arable is a separate experiment on lighter soil.
The same fitting rules keep 490 curves across 43 years. Woburn is never
used for training or tuning. It is a pure test of transfer.

\section{A profit based test bench}\label{sec:framework}

\subsection{Profit in grain terms}
We express profit in grain terms, which removes all absolute prices.
The profit of rate $N$ on a field with measured curve $Y(\cdot)$ is
\begin{equation}
  \pi_b(N) \;=\; Y(N) \;-\; \frac{b}{1000}\,N ,
  \label{eq:profit}
\end{equation}
where $b$ is the price ratio defined above. One sweep of $b$ from 3 to
12 brackets every observed UK price scenario
(Table~\ref{tab:banchors}). The observed range was 4.1 to 10.7. A piece
of advice $\hat N$ is scored by its \emph{profit loss}. That is the
profit at the best rate minus the profit at $\hat N$. Scoring always
uses curves from held out years.

\begin{table}[t]
  \centering\small
  \caption{Observed UK prices and the implied price ratio $b$. AN is
  ammonium nitrate, 34.5\,\% N. Sources: AHDB fertiliser price series;
  RB209 2022 revision reference prices.}
  \label{tab:banchors}
  \begin{tabular}{lcccc}
    \toprule
    Period & AN, \pounds/t & Wheat, \pounds/t & N price, \pounds/kg & $b$ \\
    \midrule
    2020 (Jan)        & 234 & 165 & 0.68 & 4.1 \\
    2021 (RB209 ref.) & 345 & 200 & 1.00 & 5.0 \\
    2021 (Sep)        & 395 & 200 & 1.14 & 5.7 \\
    2022 (Jul peak)   & 841 & 280 & 2.44 & 8.7 \\
    2023 (Jan)        & 700 & 190 & 2.03 & 10.7 \\
    2023 (May)        & 390 & 185 & 1.13 & 6.1 \\
    \bottomrule
  \end{tabular}
\end{table}

\subsection{Why rate error is the wrong score}
\begin{figure}[t]
  \centering
  \includegraphics[width=\textwidth]{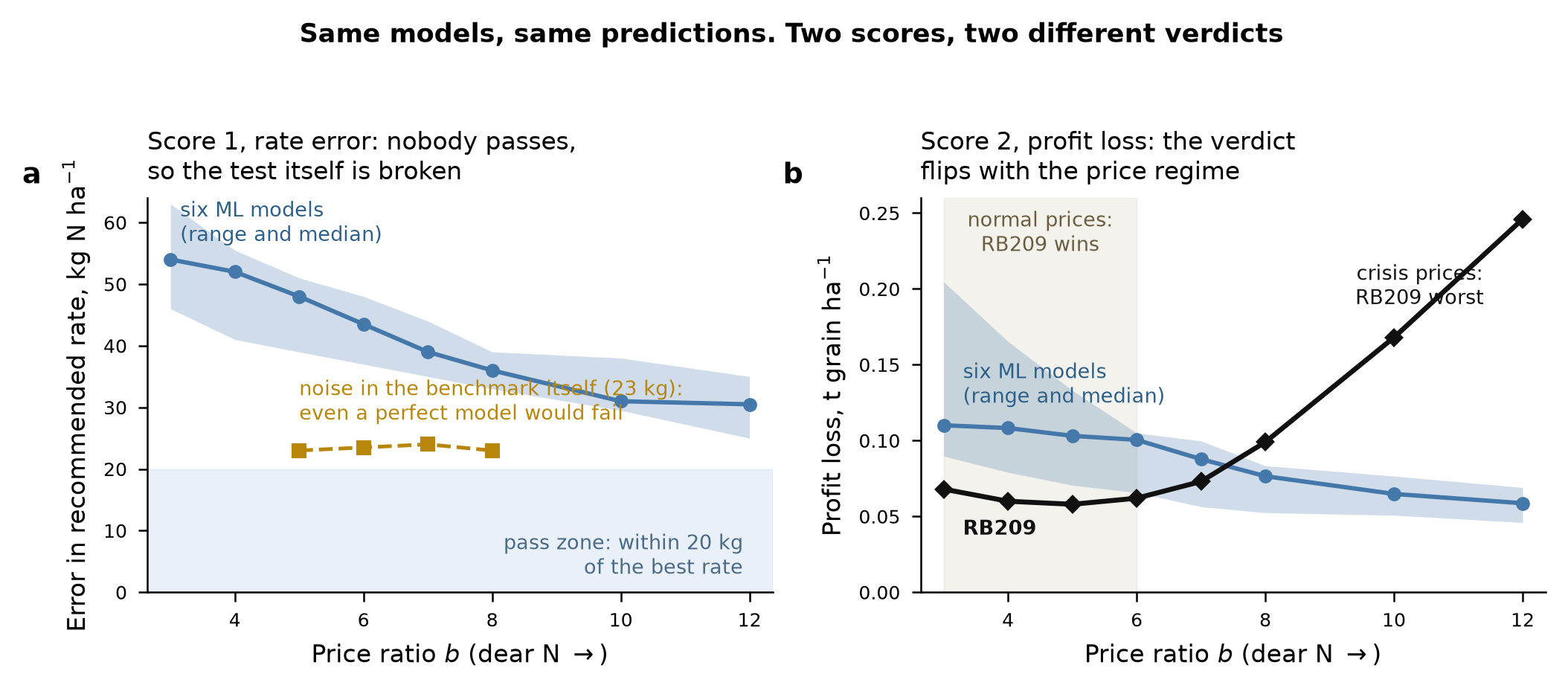}
  \caption{\textbf{Same models, same predictions, two verdicts.} The
  blue band spans our six ML models: mean curve, field mean curve,
  ridge regression, Gaussian process, random forest, and TabPFN. The
  blue line is their median. Each point covers 317 to 402 held out
  curves.
  \textbf{a}, Scored by rate error, the whole band sits above the 20~kg
  pass zone. The gold squares show the benchmark's own noise, about
  23~kg. Even a perfect model would therefore fail. The test, not the
  models, is broken.
  \textbf{b}, Scored by profit loss, the verdict depends on prices.
  RB209 (black) beats the whole band at normal prices ($b \le 6$). At
  crisis prices ($b \ge 8$) it becomes the worst option.}
  \label{fig:metric}
\end{figure}

Two facts rule out rate error as the main score. First, the benchmark
itself is noisy. Each measured curve rests on a handful of plot yields.
Refitting a curve with one observation removed moves its own best rate.
The median move is about \SI{23}{\kilo\gram\per\hectare}
(Fig.~\ref{fig:metric}a). A pass line of 20~kg would therefore fail a
perfect model. Second, profit is flat near the best rate. A 20~kg rate
error costs a median of only 0.023~\tha{} at $b=5$. A 30~kg error costs
0.052~\tha. Rate error thus punishes harmless disagreement and hides
the rare expensive misses. Profit loss prices both correctly.

\section{Machine learning alone does not clear the bar}\label{sec:negative}

This section states the negative result plainly. Six model families
were trained to predict the three curve points. They are a mean curve,
a field mean curve, ridge regression, a Gaussian process, a random
forest, and TabPFN \citep{tabpfn}. Scoring held out each of the 53
years in turn.

At normal prices, every family loses to the standard advice. For
$b \le 6$, the median profit loss of every ML family exceeds RB209's
(Fig.~\ref{fig:metric}b). This covers the price regime RB209 was built
for. The ML curves cross below RB209 only at crisis prices, $b \ge 7$.
Advice that wins only in a crisis is not yet a deployable product. This
motivates the designs of \S\ref{sec:decision}.

Unconstrained ML is worse than its parts. We let the best single
family, ridge, predict the curve points freely. Its advice loses
0.070~\tha{} at the median, double RB209's 0.035
(Table~\ref{tab:hybrid}). It also overshoots the best rate by 25~\kgN{}
on average. One byproduct is still useful. Disagreement across the six
models (median 13.6\%) sits at the low end of published values, 13.3 to
31.5\% \citep{tanaka}. Disagreement also tracks the true error
(Spearman $\rho = 0.48$), so it serves as a warning signal.

\begin{table}[t]
  \centering\small
  \caption{Three designs on held out Broadbalk curves. Numbers are
  medians over the $b$ grid; $n$ is 338 to 345 per design. Large
  errors are curves losing more than 0.3~\tha. The hybrid's median
  edge over RB209 (0.033 vs 0.035) is small and untested statistically;
  read it as ``no worse''. The hybrid's real gains are the bias and
  tail columns.}
  \label{tab:hybrid}
  \begin{tabular}{lcccc}
    \toprule
    Design & Median loss, \tha & Worst 10\% & Bias, \kgN & Large errors \\
    \midrule
    RB209 alone                     & 0.035 & 0.562 & $+6$  & 42 \\
    ML alone                        & 0.070 & 0.872 & $+25$ & 62 \\
    Hybrid: RB209 $+$ damped ML     & 0.033 & \textbf{0.518} & $\mathbf{+1}$ & \textbf{37} \\
    \bottomrule
  \end{tabular}
\end{table}

\section{The gain lives in the correction step}\label{sec:decision}

\begin{figure}[t]
  \centering
  \includegraphics[width=\textwidth]{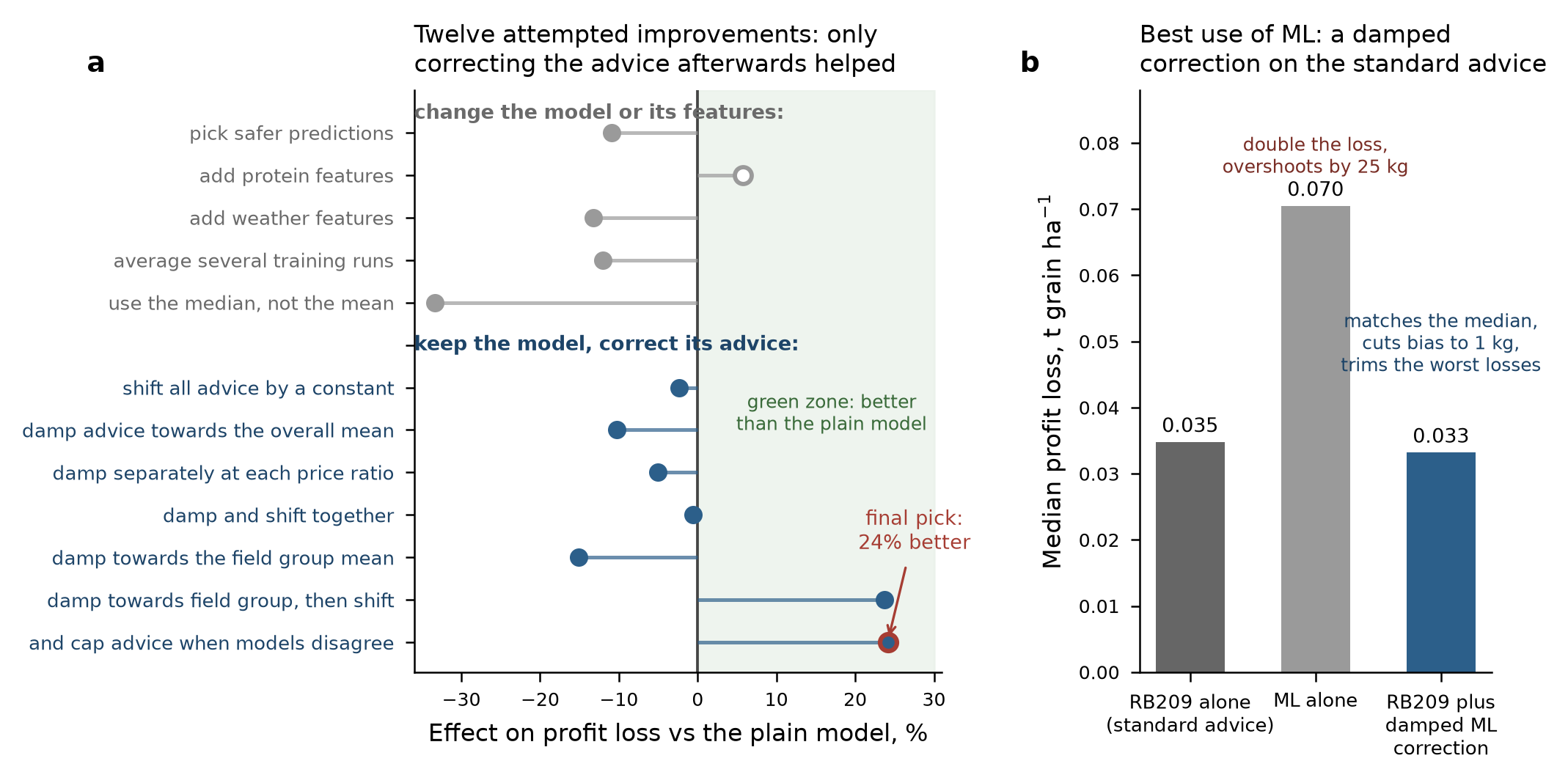}
  \caption{\textbf{Where the gain comes from.}
  \textbf{a}, We tried twelve improvements to one ML pipeline. Each was
  scored with every year held out in turn. Bars to the right mean lower
  profit loss than the plain model. Grey group: better models and extra
  features; none helped reliably. Blue group: keep the model and
  correct its advice afterwards; only these reached the green zone. The
  open ring improved here but not in the original run.
  \textbf{b}, The three way comparison of Table~\ref{tab:hybrid}. ML
  alone doubles the loss of the standard advice. Adding a damped ML
  correction to RB209 matches its median. It also removes bias and
  trims the worst losses.}
  \label{fig:decision}
\end{figure}

\subsection{The exploration ledger}
We tested twelve pipeline changes under one locked protocol. The
baseline is TabPFN with an evaluation loss of 0.080~\tha. The protocol
held out every year in turn. Truth tables were locked by checksum, and
the evaluator ran separately from the candidate author.
Table~\ref{tab:ledger} lists every change and its effect. The changes
fall into two groups. One group changes the model or its features. The
other group leaves the model alone. It corrects the model's advice
afterwards, using three simple tools. \emph{Damping} pulls each advice
part way towards a group average. \emph{Shifting} adds one constant
offset, learned from training years. \emph{Capping} limits the
correction where the six models disagree.

\begin{table}[t]
  \centering\small
  \caption{The twelve attempted improvements, in the order of
  Fig.~\ref{fig:decision}a. Effect is the change in profit loss against
  the plain model. Positive means better.}
  \label{tab:ledger}
  \begin{tabular}{p{4.6cm}p{7.0cm}r}
    \toprule
    Change tested & What the operation does & Effect \\
    \midrule
    \multicolumn{3}{l}{\emph{Change the model or its features}} \\
    Pick safer predictions & Choose the rate with the smallest
      predicted downside, not the highest mean profit. & $-10.9$\,\% \\
    Add protein features & Add grain protein measurements as extra
      model inputs. & $+5.7$\,\% \\
    Add weather features & Add season rainfall, temperature, and
      radiation as extra inputs. & $-13.2$\,\% \\
    Average several training runs & Train the model several times with
      different seeds; average the predictions. & $-12.0$\,\% \\
    Use the median, not the mean & Combine those runs by median instead
      of mean. & $-33.4$\,\% \\
    \midrule
    \multicolumn{3}{l}{\emph{Keep the model, correct its advice}} \\
    Shift all advice by a constant & Learn one offset from training
      years; add it to every advice. & $-2.3$\,\% \\
    Damp towards the overall mean & Pull each advice part way towards
      the average advice. & $-10.3$\,\% \\
    Damp separately at each price ratio & As above, with the damping
      strength tuned per price ratio. & $-5.1$\,\% \\
    Damp and shift together & Combine the damping and the constant
      shift. & $-0.6$\,\% \\
    Damp towards the field group mean & Pull each advice towards the
      average of its own field group. & $-15.1$\,\% \\
    Damp towards field group, then shift & Field group damping plus the
      constant shift. & $+23.7$\,\% \\
    Cap advice when models disagree & Where the six models disagree
      widely, limit the correction. The final pick. & $+24.1$\,\% \\
    \bottomrule
  \end{tabular}
\end{table}

\subsection{Results, stated with care}\label{sec:significance}
The whole gain came from the correction group.
Figure~\ref{fig:decision}a shows the ledger. Model and feature changes
gave no reliable gain. The final correction pick scores 0.061, which is
24\% below baseline. Three qualifications matter. First, the planned
mean tests do not reach significance (Wilcoxon $p=0.70$), because a few
extreme years dominate yearly means. Second, a bootstrap over years
gives $\Delta = 0.019$ with CI $[-0.001, 0.057]$ and one sided
$p = 0.037$. Third, one feature change (open ring) improved here but
not in the original run. We report that reversal rather than hide it.
Our claim is therefore narrow. The largest gain, and the only gain that
transfers, is the correction step. The strongest evidence is the
transfer test below.

One misreading should be closed off. The correction does not make the
model redundant. The tuned damping pulls each advice only part way
towards its group average. Damping all the way would discard the model
entirely. That variant scored 15\% worse than the plain model
(Table~\ref{tab:ledger}). The model therefore supplies real field to
field differences worth keeping. The uncertainty cap also needs the
model ensemble, since it reads their disagreement. The model must be
good enough to capture the curve's slope. Beyond that point, further
modelling effort stopped paying; using its output well did.

\section{The correction transfers to a new site}\label{sec:external}

\begin{figure}[t]
  \centering
  \includegraphics[width=0.55\textwidth]{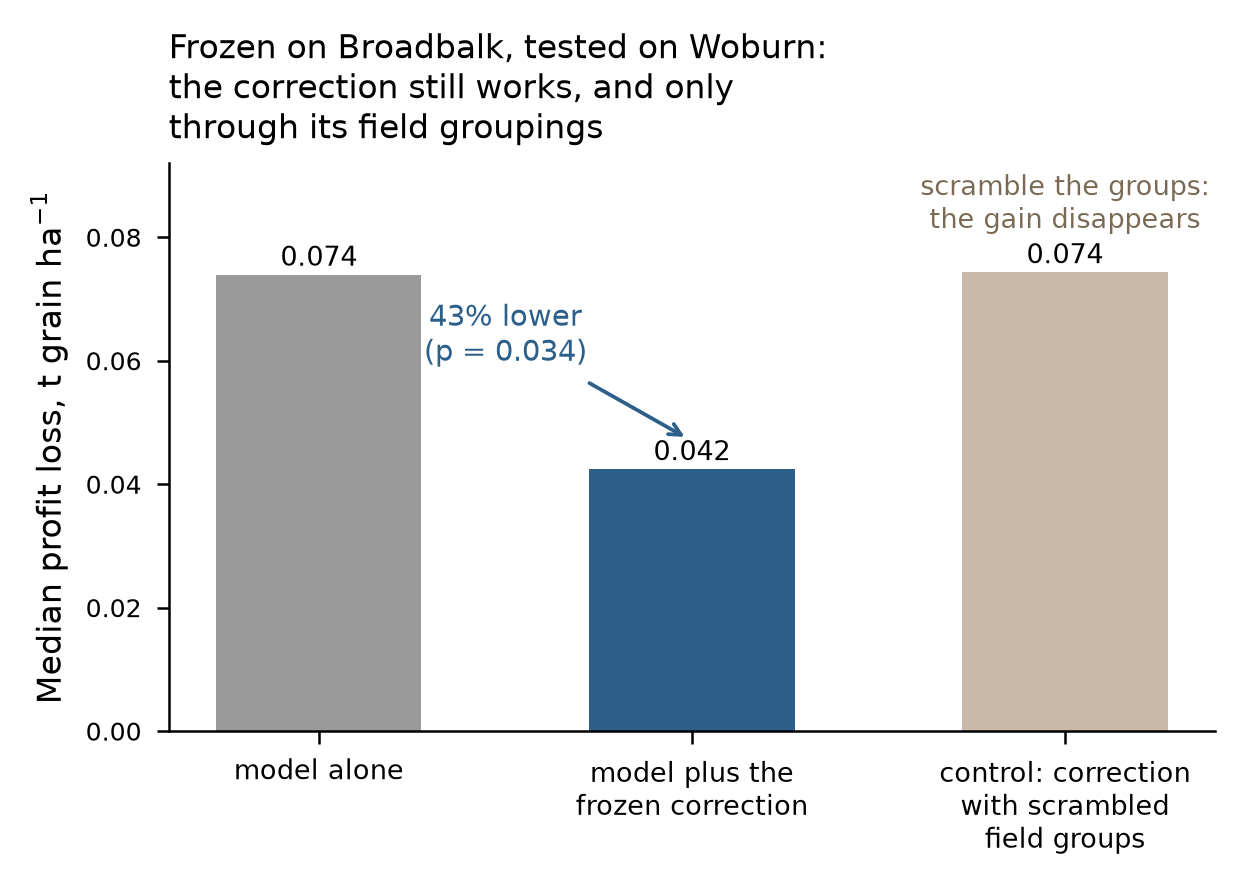}
  \caption{\textbf{Transfer with no retraining.} The pipeline was frozen
  on Broadbalk and applied to 490 Woburn curves unchanged. Middle bar:
  the correction cuts the median profit loss by 43\% ($p=0.034$). Right
  bar, the control: scramble which field group each curve belongs to,
  keeping everything else. The whole gain disappears. So the correction
  works through real field groupings, not luck.}
  \label{fig:external}
\end{figure}

The frozen correction works at a site it never saw. Woburn differs in
soil, rotations, and experimental design. Nothing was retrained or
tuned there. The frozen pipeline cut the median loss from 0.074 to
0.042~\tha, a 43\% drop (Wilcoxon $p = 0.034$). The original study
reported 42\% on the same test.

A control confirms the mechanism. Recall how the correction works: it
pulls each curve's advice towards its field group average. The control
repeats the whole test with one change. We deliberately mismatch the
curves and the group labels. Each advice is then pulled towards the
average of the wrong group. Everything else stays identical: same
model, same damping, same shift. If the gain were a generic averaging
effect, it would survive this. It does not. The loss returns to the
uncorrected level (0.074, Fig.~\ref{fig:external}, right bar). So the
field groups carry real agronomic information, and the correction
works through it. This transfer is the strongest single piece of
evidence in the paper. It is independent of the data that selected the
method.

An architectural point explains why correction beats prediction, and
it is the same decoupling shown in Fig.~\ref{fig:teaser}. The advice
depends on the slope of the yield curve, not its height. Adding a
constant to every yield prediction changes no advice. A correction step
targets the slope; better prediction mostly fixes the height.

\section{Profit, prices, and emissions on one dial}\label{sec:economics}

\begin{figure}[t]
  \centering
  \includegraphics[width=0.6\textwidth]{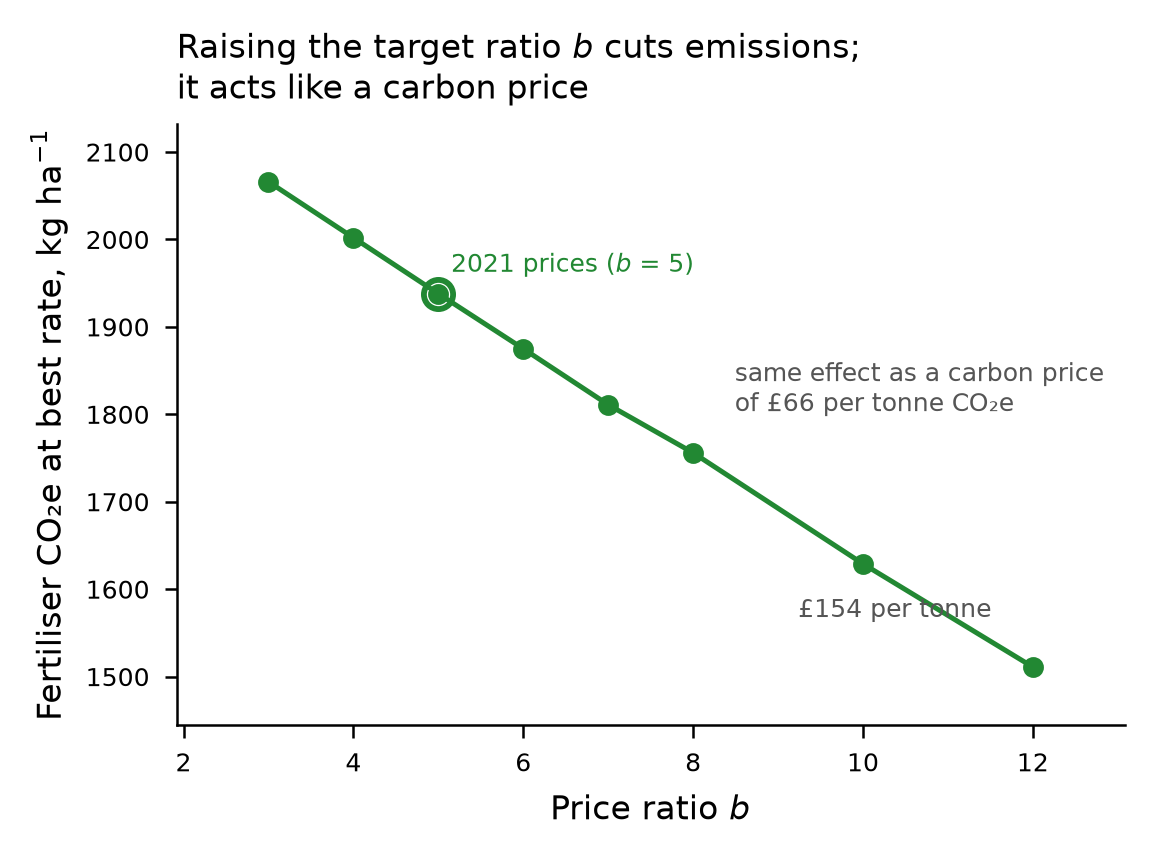}
  \caption{\textbf{The price ratio is also a carbon dial.} Fertiliser
  emissions at the best measured rate, across the price ratio $b$. The
  factor is \SI{9.10}{\kilo\gram}~CO$_2$e per kg N: manufacture 3.42
  \citep{brentrup2018} plus field nitrous oxide 5.68
  \citep{ipcc2019,ipccar6}. Raising the target from $b=5$ to $b=8$ has
  the same effect as a \pounds 66 per tonne carbon price and saves
  about \SI{180}{\kilo\gram}~CO$_2$e per hectare.}
  \label{fig:slider}
\end{figure}

\subsection{What the flat profit curve means}
The economic case does not live in the median gain. Profit is flat near
the best rate, so the median gain of any advice over RB209 is single
digit pounds per hectare (Table~\ref{tab:gbp}). The case rests on three
legs instead.
\begin{enumerate}
  \item \textbf{Rare large losses.} RB209's worst 10\% of losses reach
        1.7 to 3.5~\tha, depending on $b$. That is \pounds 275 to 970
        per hectare. The hybrid trims the count of large errors from 42
        to 37 while its median stays level with RB209. The insurance
        comes at no premium.
  \item \textbf{Price response.} At 2023 prices ($b \approx 10.7$), the
        best rate sits about 38~\kgN{} below the $b=5$ rate. Following
        that shift saves about \pounds 82 per hectare of fertiliser
        spend. Profit barely moves, because the curve is flat. Fixed
        advice captures none of this.
  \item \textbf{Emissions.} The same downshift cuts fertiliser
        emissions. The cost per tonne saved is quantified below.
\end{enumerate}

\begin{table}[t]
  \centering\small
  \caption{Headline results converted to \pounds{} per hectare. The
  conversion multiplies grain terms by the grain price.}
  \label{tab:gbp}
  \begin{tabular}{lccccc}
    \toprule
    Quantity & \tha{} grain & @160 & @200 & @240 & @280 \\
    \midrule
    RB209 median loss, $b{=}5$    & 0.058 & 9   & 12  & 14  & 16 \\
    RB209 median loss, $b{=}12$   & 0.246 & 39  & 49  & 59  & 69 \\
    RB209 worst 10\%, $b{=}3$     & 3.45  & 553 & 691 & 829 & 967 \\
    Hybrid vs ML alone            & 0.037 & 6   & 7   & 9   & 10 \\
    Correction step gain          & 0.019 & 3   & 4   & 5   & 5  \\
    Woburn transfer gain          & 0.032 & 5   & 6   & 8   & 9  \\
    \bottomrule
  \end{tabular}
\end{table}

\subsection{A worked carbon cost}
The price ratio converts directly into a carbon price. Each kilogram of
fertiliser N carries about \SI{9.10}{\kilo\gram}~CO$_2$e. Manufacture
contributes 3.42 for modern European ammonium nitrate
\citep{brentrup2018}. Field nitrous oxide contributes 5.68 under IPCC
default factors \citep{ipcc2019}, using the AR6 warming value of 273
\citep{ipccar6}. A carbon price $c$ on these emissions raises the
effective N price by $c$ times 9.10 divided by 1000. At grain
\pounds 200 per tonne, \pounds 66 per tonne CO$_2$e moves $b$ from 5 to
8, and \pounds 154 moves it to 12 (Fig.~\ref{fig:slider}).

The measured curves then price the emission cut for the farmer. We take
true prices at $b=5$ and follow the $b=8$ advice. The median across 335
curves loses 0.039~\tha, or \pounds 7.9 per hectare at \pounds 200
grain. It saves about \SI{246}{\kilo\gram}~CO$_2$e per hectare. The
cost is \pounds 32 per tonne of CO$_2$e saved. The deeper cut to
$b=12$ costs \pounds 72 per tonne. Both sit at or below recent UK
carbon price levels. No separate optimiser is needed. Emission cuts are
a position on the price dial the model already has.

\subsection{Scale}
The stakes are national. UK winter wheat covers about 1.6 million
hectares. The price response leg alone is worth order \pounds 100M in a
high price year.

\section{Limitations}\label{sec:limitations}
\begin{itemize}
  \item Training data come from one site, Broadbalk. Woburn tests
        transfer; it is not a second training site.
  \item The hybrid's median edge over RB209 is untested statistically.
        Read it as ``no worse''. Its case rests on bias and tail risk.
  \item The correction gain fails the planned mean tests at 53 years.
        The bootstrap interval just touches zero in our rerun. The
        Woburn transfer ($p=0.034$) carries the statistical weight.
  \item One ledger entry reversed direction between runs, and six of
        thirteen moved. The final ranking and the main conclusion held
        in every rerun.
  \item Emission factors are defaults, not field measurements.
  \item Grain terms omit application costs, protein premiums, and
        storage. The pound conversions are rescalings, not farm
        budgets.
\end{itemize}

\section{Conclusion}\label{sec:conclusion}
Does machine learning pay for nitrogen advice? Posed as prediction, no.
Every model failed on rate error, and the benchmark noise shows any
model would. At normal prices, every model also lost to the standard
advice on profit. Posed as a decision aid, yes, under three conditions.
Score it in profit, on measured curves. Use it as a damped correction
on the standard advice. Sweep the price ratio, so the advice responds
to prices. Under these conditions the gains are real. Bias falls from 6
to 1~\kgN. Rare large losses shrink. The correction transfers to a new
site unchanged, 43\% better. The same price dial cuts emissions at
\pounds 32 to 72 per tonne. The answer is a conditional yes.

\section*{Reproducibility}
The evaluation used locked truth tables and held out years throughout.
Curve tables, ledgers, per curve losses, and scripts will be released.

\section*{Data availability}
Both field datasets are open access under a CC BY 4.0 licence.
Broadbalk wheat yields, 1968 to 2022: \citet{bbkdata},
doi:10.23637/rbk1-yld6822-01. Woburn Ley arable wheat yields, 1976 to
2018: \citet{wrndata}, doi:10.23637/wrn3-wheat7618-01. Both are
published by the electronic Rothamsted Archive (e-RA), Rothamsted
Research, Harpenden, UK.

\section*{Acknowledgements}
We thank Rothamsted Research and the e-RA curators for access to the
Broadbalk and Woburn Ley arable data. The Rothamsted Long Term
Experiments National Bioscience Research Infrastructure is supported
by UKRI-BBSRC and the Lawes Agricultural Trust.

\bibliographystyle{plainnat}

\begin{thebibliography}{9}\small
\bibitem[AHDB(2023)]{ahdb2022} AHDB. \emph{Nutrient Management Guide
(RB209), Section 4: Arable crops.} Agriculture and Horticulture
Development Board, 2023 edition; economic adjustment tables revised 2022.
\bibitem[Tanaka et~al.(2024)]{tanaka} Tanaka, T.\,S.\,T., Heuvelink,
G.\,B.\,M., Mieno, T., and Bullock, D.\,S. Can machine learning models
provide accurate fertilizer recommendations? \emph{Precision
Agriculture} 25:1839--1856, 2024. doi:10.1007/s11119-024-10136-x.
\bibitem[IPCC(2019)]{ipcc2019} IPCC. \emph{2019 Refinement to the 2006
IPCC Guidelines for National Greenhouse Gas Inventories}, Vol.~4,
Ch.~11: N$_2$O emissions from managed soils. IPCC, 2019.
\bibitem[IPCC(2021)]{ipccar6} IPCC. \emph{Climate Change 2021: The
Physical Science Basis (AR6 WG1)}, Ch.~7 Supplementary Material,
Table~7.SM.7. IPCC, 2021.
\bibitem[Brentrup et~al.(2018)]{brentrup2018} Brentrup, F., Lammel, J.,
Stephani, T., and Christensen, B. Updated carbon footprint values for
mineral fertilizer from different world regions. In \emph{Proc.\ 11th
Int.\ Conf.\ on Life Cycle Assessment of Food}, 2018.
\bibitem[Hollmann et~al.(2025)]{tabpfn} Hollmann, N., M\"uller, S.,
Purucker, L., et~al. Accurate predictions on small data with a tabular
foundation model. \emph{Nature} 637:319--326, 2025.
\bibitem[Glendining and Poulton(2023)]{bbkdata} Glendining, M. and
Poulton, P. Dataset: Broadbalk Wheat annual grain and straw yields
1968--2022. Electronic Rothamsted Archive, Rothamsted Research,
Harpenden, UK, 2023. doi:10.23637/rbk1-yld6822-01.
\bibitem[Glendining et~al.(2022)]{wrndata} Glendining, M., Poulton, P.,
Macdonald, A., MacLaren, C., and Clark, S. Dataset: Woburn Ley-arable
experiment: yields of wheat as first test crop, 1976--2018. Electronic
Rothamsted Archive, Rothamsted Research, Harpenden, UK, 2022.
doi:10.23637/wrn3-wheat7618-01.
\end{thebibliography}

\end{document}